\documentclass[11pt]{article}
\usepackage[final]{acl}
\usepackage{times}
\usepackage{latexsym}
\usepackage[T1]{fontenc}
\usepackage[utf8]{inputenc}
\usepackage{microtype}
\usepackage{inconsolata}
\usepackage{graphicx}
\usepackage{booktabs}
\usepackage{enumitem}
\usepackage{tabularx}
\usepackage{adjustbox}
\usepackage{float}
\usepackage{xcolor}
\usepackage{placeins}
\usepackage{pifont}
\usepackage{hyperref}
\usepackage{amsmath}
\usepackage{amsfonts}
\usepackage{inconsolata}
\usepackage{fontawesome}
\usepackage{graphicx}
\usepackage{url}  
\usepackage{caption}
\usepackage[table]{xcolor}
\usepackage{color}

\definecolor{oursrow}{RGB}{255,235,235}

\title{Small yet Assistive: Spatially-Aware Post-Training for Low Vision}

\author{
  Rishabh Choudhary$^{1*}$, 
  Shreyansh Raj$^{1*}$, 
  Umesh Goyal$^{1}$,
  Shubh Kashyap$^{1}$, 
  Shrestha Kumar$^{1}$, \\
  \textbf{Sushovan Jena}$^{1*\dagger}$,
  \textbf{Komal Kumar}$^{2*}$,
  \textbf{Hisham Cholakkal}$^{2}$,
  \textbf{Aditya Nigam}$^{1}$\\
  \textsuperscript{1}Indian Institute of Technology Mandi\\
  $^{2}$ Mohamed bin Zayed University of Artificial Intelligence \\
\small
\begin{tabular}{cc}
{\fontsize{10}{10}\selectfont\faGithub}~\textbf{GitHub:}~\href{https://github.com/Shreyansh262/Small-Yet-Assistive}{\texttt{\textcolor{teal}{github.com/Shreyansh262/Small-Yet-Assistive}}}
\end{tabular} \\[2pt]
\small
{\fontsize{10}{10}\selectfont\faGlobe}~\textbf{Website:}~\href{https://smol-vl-blv.github.io/Smol-VL-BLV-website/}{\texttt{\textcolor{teal}{smol-vl-blv.github.io/Smol-VL-BLV-website/}}}
}

\begin{document}
\makeatletter
\let\@oldmaketitle\@maketitle
\renewcommand{\@maketitle}{%
  \@oldmaketitle
  \begin{center}
    \includegraphics[width=\textwidth,height=0.5\textheight,keepaspectratio]{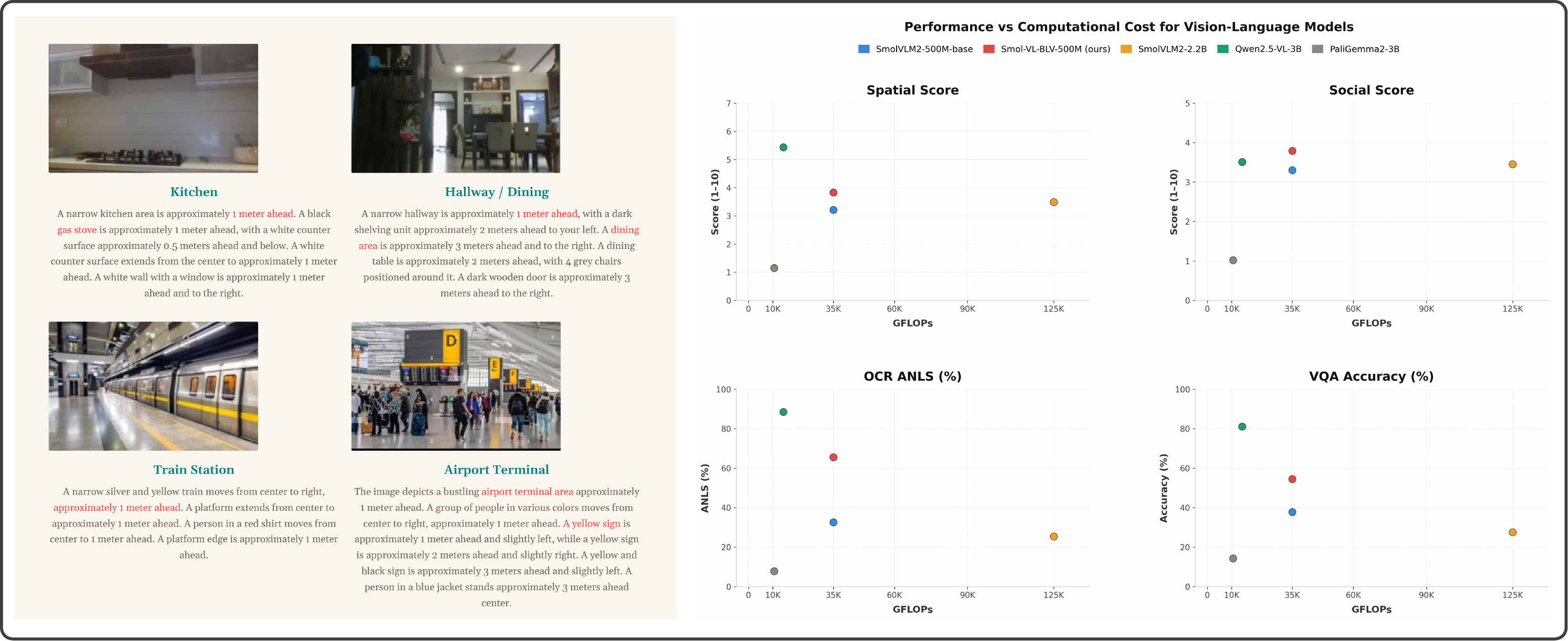}
    \captionof{figure}{Overview of our on-device BLV-oriented vision-language model. 
Left: qualitative examples of generated scene descriptions across indoor, transit, and public-space settings. The highlighted phrases show BLV-critical information captured by the model, including distance estimates, directional cues, obstacles, people, signs, and spatial layout. 
Right: performance--efficiency comparison across VLMs. After spatial-grounding supervised fine-tuning and GRPO-based preference optimization, our SmolVLM-500M model improves spatial reasoning, social understanding, OCR, and VQA accuracy while remaining substantially cheaper than larger VLMs in computational cost. These results suggest that high-quality BLV-focused synthetic supervision from a stronger teacher model can induce emergent spatial grounding and more precise accessibility-oriented descriptions in compact VLMs.}
    \label{fig:main_demo}
  \end{center}
}
\makeatother
\maketitle
\let\thefootnote\relax\footnotetext{$^{*}$ Equal contribution. $^{\dagger}$ Corresponding author: \texttt{sushovanjena@gmail.com}}
\begin{abstract}
An estimated 1 billion people worldwide live with vision impairment, yet current vision-language models (VLMs) produce descriptions too vague for safe navigation by blind and low-vision (BLV) users. Large VLMs can generate high-quality audio-description-compliant narrations but cannot run on mobile devices; small VLMs offer competitive latency but lack spatial detail, directional cues, and hazard awareness for navigational assistance. We present Smol-VL-BLV, a compact VLM for blind and low-vision users that closes this gap using a 500M decoder transformer model and two post-training mechanisms: (1) teacher-student distillation and (2) Group Relative Policy Optimization (GRPO) with a composite BLV reward targeting directional language, metric distances, and hazard detection. Because multi-stage post-training can induce catastrophic forgetting, we add a lightweight finetuning stage after the last stage GRPO finetuning to recover general descriptive quality while preserving BLV-specific spatial grounding. Our best model substantially outperforms the baseline across various benchmarks, including tasks: VQA, BLV captioning, OCR, and latency. Compared with the baseline for relative improvement, it improves the Spatial score gain of 19.3\%, and the Social score gain of 14.8\%.  It also increases OCR-Bench by 101.5\%, and raises TextVQA accuracy by 44.2\%. These results show that BLV-focused post-training improves both accessibility-specific spatial grounding and general visual-text reasoning. Deployed on a mid-range Android smartphone via Mixed-Precision Quantization, the model remains approx. 450 MB and runs entirely on-device, offline and without network dependency, generating descriptions with latency dependent on host hardware capabilities. Our model, dataset, and code is publicly released at \url{https://smol-vl-blv.github.io/Smol-VL-BLV-website/}.


\end{abstract}

\section{Introduction}
\label{sec:intro}

The affected population worldwide with vision impairment\footnote{\href{https://www.who.int/news-room/fact-sheets/detail/blindness-and-visual-impairment}{blindness-and-visual-impairment}} is estimated around 1 billion which includes 94 million with cataracts and 88.4 million with uncorrected refractive errors \cite{who2024}. While smartphone-based tools such as Apple VoiceOver and Google TalkBack effectively describe user interfaces, they fall short at describing real-world scenes with the spatial detail, directional cues, and hazard awareness that blind and low-vision (BLV)~\cite{corn2010foundations,leat1999low} users require for safe, independent navigation.

Professional audio description (AD) standards, established by the ITC and adopted by Netflix~\cite{lopez2023audio} and the ACB, prescribe a specific style of visual narration: present-tense, spatially grounded, hazard-first, and objective~\cite{conway2020audio}. A description following these guidelines might read: \textit{``A corridor extends approximately eight meters ahead. A partially open door is on your left, three meters away. A person walks toward you from the far end.''} Current vision-language models (VLMs), trained predominantly on internet-scale data written for sighted audiences~\cite{wang2025scaling}, instead produce vague outputs such as \textit{``A hallway with some doors and people''}---adequate for image captioning benchmarks but inadequate for physical navigation~\cite{ourpaper2025}.


Earlier BLV assistive technology research shows a shift from basic OCR and object recognition toward richer visual access, including video description, visual question answering, spatial exploration, customizable assistance, and conversational tools~\cite{li2025videoa11y,cheema2025describe,cheema2025describepro,cheema2026vidscribe,xu2025branch,xu2026sonic,li2026adcanvas}. In parallel, efficient vision-language model research has advanced compact architectures and mobile deployment through smaller VLMs, faster vision encoding, token reduction, and low-latency inference~\cite{wang2020minivlm,marafioti2025smolvlm,chu2023mobilevlm,vasu2025fastvlm,huang2025litevlm}. However, these directions rarely combine real-time mobile capability with BLV-specific spatial grounding, hazard awareness, and concise navigation-oriented descriptions.

Prior work \cite{li2025videoa11y,ourpaper2025} benchmarked SmolVLM2-500M on a smartphone and established two domain-specific evaluation frameworks---the Multi-Context Framework (MCF) and the Navigational Assistance Framework (NAF): showing that the base model produces descriptions that are too vague and spatially unaware for BLV use. The present work closes this quality gap by developing a Smol-VL-BLV model, which consists mainly of two mechanisms: (1) \textbf{Teacher-student distillation.} We use Gemma-4-31B-IT as a teacher to generate AD-guideline-compliant captions for 9,646 videos from the Charades and AVCaps datasets. (2) \textbf{Reinforcement learning via GRPO.} We apply GRPO with a domain-specific reward targeting directional content, spatial language, and hazard mentions.

Beyond post-training, we perform a systematic deployment study targeting real-world mobile inference on a mid-range Android smartphone. Our final trained model is exported to GGUF format~\cite{tripathi2024gguf}, producing a 783 MB float16 checkpoint. To compress this for on-device use, we apply IQ4\_NL quantization guided by an importance matrix computed over our Gemma-generated BLV navigation captions: the same captions used during training.  Within this scheme, the attention layers that govern where the model focuses spatially are kept at a higher precision (Q5\_K) than the rest of the network, and the vision projector is held at float16 throughout. The result is a 442MB two-file deployment package: a 251MB quantized language model backbone and a 191MB vision projector: that runs via CPU-only inference on a Samsung A55, achieving 39.3 tok/s generation throughput and 27.1s end-to-end latency per frame.

We conduct comprehensive experiments using the Charades and AVCaps datasets, evaluating across keyword-based BLV and navigational scoring, standard NLP metrics, and LLM-as-judge assessment. Our model achieves strong results on spatial orientation, social interaction, navigational coverage, and general vision-language benchmarks, including OCR and VQA. Figure~\ref{fig:main_demo} illustrates qualitative examples and a cross-model performance comparison. Our main contributions are:
\begin{itemize}
    \item A complete teacher-student pipeline for distilling BLV-compliant video description capability into a $<1B$ VLM deployable on commodity smartphones.
    \item A multi-stage post-training strategy combining SFT, GRPO, and patch-based refinement, improving BLV coverage, spatial grounding, navigational cues, and OCR/VQA transfer while yielding the best overall model across evaluation metrics.
    \item Comprehensive evaluation across BLV-specific metrics, OCR, VQA,  and on-device benchmarking demonstrating both model quality and practical viability for real-world BLV assistance.
\end{itemize}
\section{Related Work}
\paragraph{BLV-Centered Visual Assistance.}
Assistive technologies for BLV users help them access visual information through OCR readers, object recognition tools, navigation aids, wearable cameras, screen readers, smartphone apps, and remote human assistance. However, BLV users do not all need the same kind of help. Their needs depend on the task, environment, level of vision, and type of information required~\cite{corn2010foundations,leat1999low,who2024}. Recent systems move beyond only detecting objects or reading text by supporting video description, visual question answering, spatial exploration, customizable descriptions, and conversational assistance~\cite{li2025videoa11y,cheema2025describe,cheema2025describepro,cheema2026vidscribe}. Other works such as Branch Explorer, Sonic Stage, and ADCanvas show the importance of spatial understanding, interaction, and user control~\cite{xu2025branch,xu2026sonic,li2026adcanvas}. Still, many systems focus on video/media access or authoring workflows rather than real-time scene understanding for navigation and hazard awareness. This motivates our focus on short, spatially grounded, and safety-aware BLV descriptions instead of generic captions.
\paragraph{Real-Time \& On-Device Multimodal Systems.}
In the development of real-time and on-device multimodal systems, vision-language models are designed to work efficiently on mobile devices, edge devices, and other resource-constrained environments. This is especially important for BLV assistance, because delayed responses may be unsafe or unhelpful during navigation, obstacle detection, or hazard avoidance. Prior work on efficient VLMs has focused on reducing model size, improving vision encoders, compressing visual tokens, using speculative decoding, and applying quantization to reduce memory and computation cost~\cite{chu2023mobilevlm,vasu2025fastvlm,huang2025litevlm,marafioti2025smolvlm}. Recent work on lightweight VLMs for BLV accessibility also shows the need to evaluate compact models for accessibility-specific use cases rather than only on generic benchmarks~\cite{ourpaper2025}. However, most prior work mainly optimizes general-purpose VLM efficiency and does not directly address BLV-specific requirements such as when to warn users, when to express uncertainty, how to describe hazards, or how to provide concise spatial guidance. Our work builds on efficient lightweight VLM research, but specifically targets low-latency, on-device BLV visual assistance by adapting a compact model for spatially grounded and hazard-aware descriptions.


\section{Methodology}
\label{sec:method}

\begin{figure*}[t]
\centering
\includegraphics[width=\textwidth]{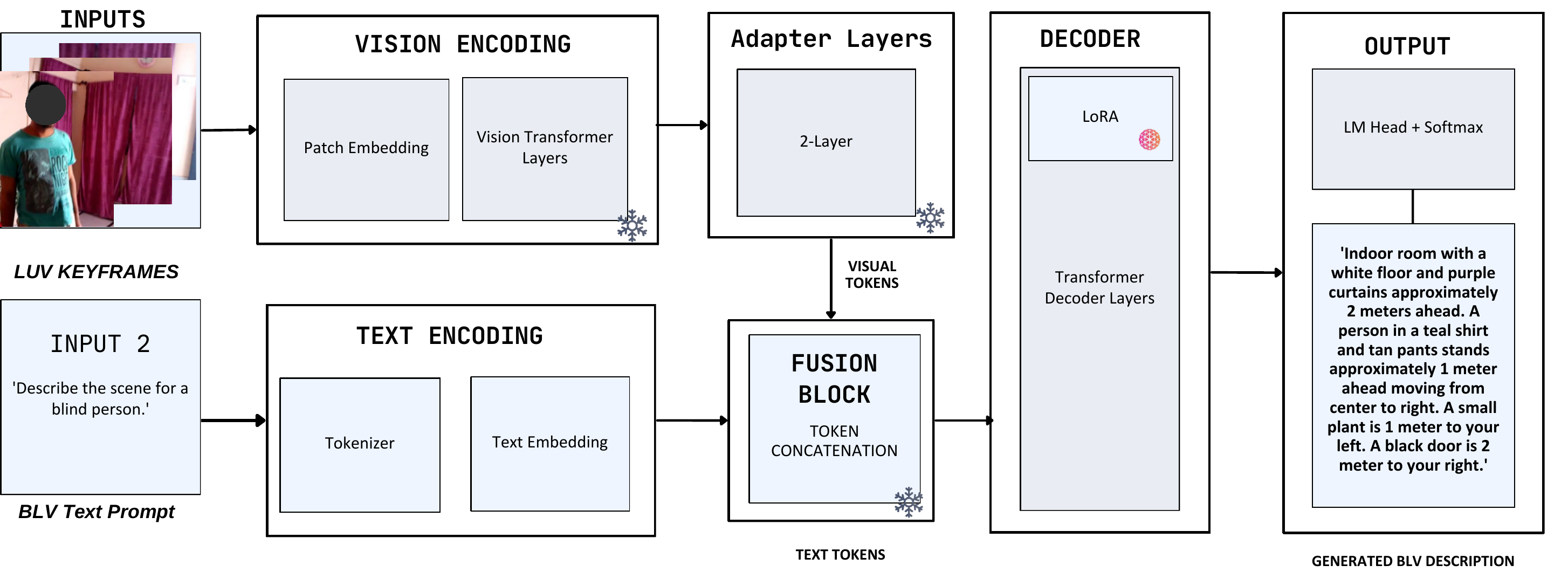}
\caption{
Overview of the post-training pipeline. LUV keyframes and a BLV-oriented text prompt are encoded separately through the vision and text encoders, aligned through lightweight adapter layers, and merged in a fusion block via token concatenation. The decoder, fine-tuned with LoRA while keeping selected backbone components frozen, generates spatially grounded BLV descriptions that include distances, directions, objects, obstacles, etc.
}
\label{fig:training_pipeline}
\end{figure*}

Our pipeline comprises four stages: (1) data preparation—keyframe extraction and teacher caption generation; (2) supervised fine-tuning with phase-switching (SFTv2); (3) reinforcement learning via GRPO with a BLV-specific reward; and (4) SFT-patch recovery and mobile deployment. Figure\ref{fig:training_pipeline} provides an overview. The student model throughout is SmolVLM2-500M-Video-Instruct (Marafioti et al., 2025b): a SigLIP-based vision tower (86.4M), a multi-modal connector ( 11.8M), and a LLaMA-architecture (SmolLM2) language decoder with LM head ( 409.3M), totalling  507.5M parameters. Following the SmolVLM naming convention, "500M" is used throughout the paper as a colloquial class label rather than a literal parameter sum.

\subsection{Data Preparation}
\label{sec:data_prep}

\paragraph{Keyframe Extraction}
\label{sec:keyframes}

We extract $n{=}4$ keyframes per video using perceptual color-space differencing in LUV space. After subsampling to at most 64 evenly-spaced frames, we compute the mean absolute inter-frame difference:
\begin{equation}
    d_i = \frac{1}{HW} \sum_{h,w} \left| \mathrm{LUV}(f_i)_{h,w} - \mathrm{LUV}(f_{i-1})_{h,w} \right|
    \label{eq:luv}
\end{equation}
The $n{-}1$ highest-$d_i$ frames plus $f_1$ are selected and sorted chronologically, capturing scene transitions and significant motion events. LUV differencing requires no additional model inference—unlike DINOv2\cite{oquab2023dinov2} semantic embeddings (86M parameters, 50--100\,ms overhead)—making it suitable for mobile deployment.

\begin{figure}[t!]
    \centering
    \includegraphics[width=\linewidth]{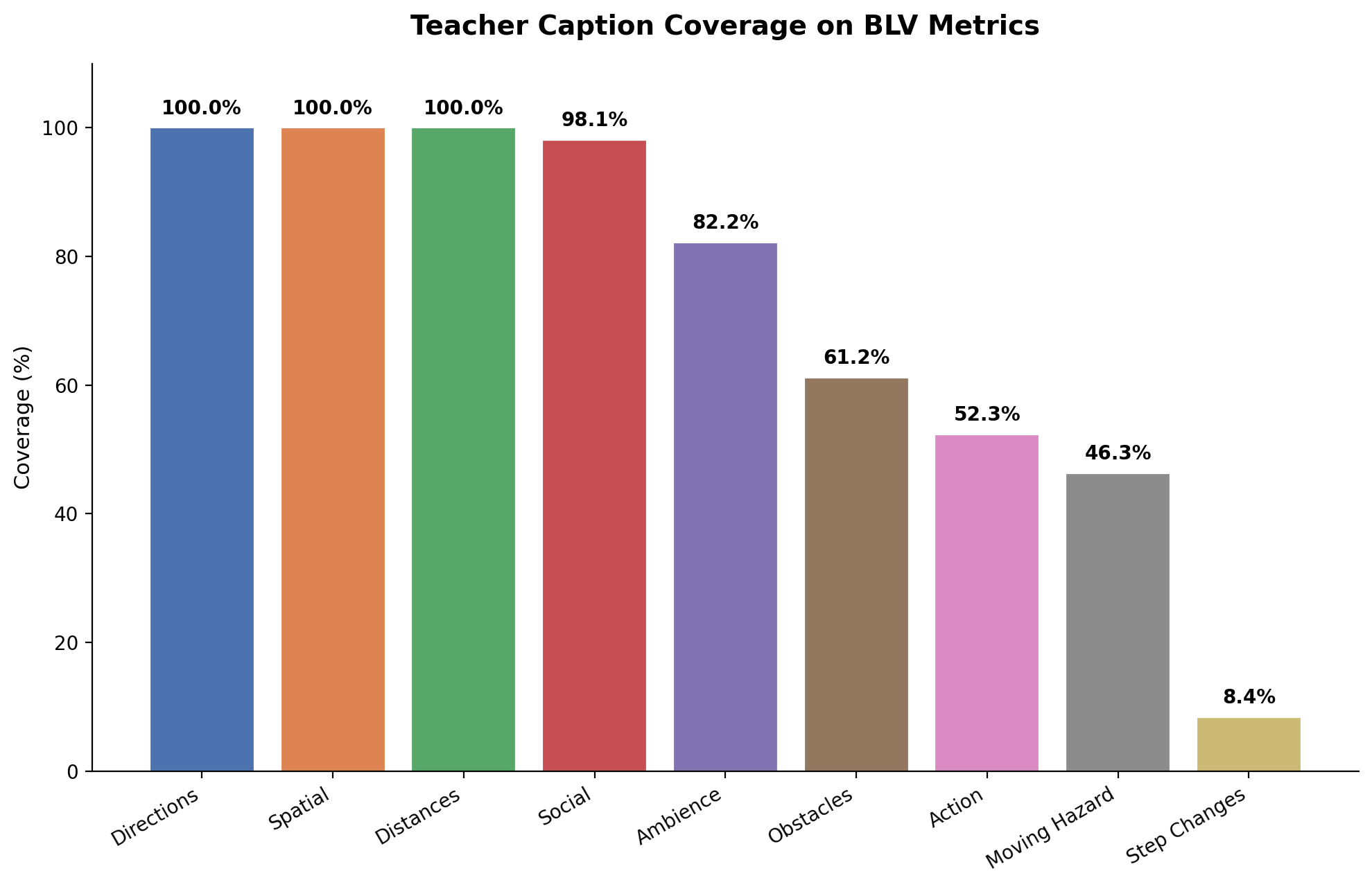}
    \caption{
Coverage of BLV-relevant attributes in teacher-generated captions. The captions consistently include core navigation cues such as directions, distances, and spatial layout, while also covering social context, ambience, obstacles, actions, and moving hazards. Lower coverage for step changes indicates an important remaining challenge for BLV-oriented scene description.
}

    \label{fig:data_chart}
\end{figure}

\paragraph{Teacher Caption Generation}
\label{sec:teacher}
We use Gemma-4-31B-IT\footnote{\href{https://huggingface.co/google/gemma-4-31B-it}{Gemma-4-31B-IT}} as the teacher, as it is a very powerful image-to-text model and shows excellent detailing and grounding in video. The system prompt encodes professional AD guidelines \cite{bittner2012audio}: present tense, safety-critical content first, explicit spatial positions with distances, observable features only, environment type in the first sentence, hazard/obstacle/level-change flags, active voice, maximum 4 sentences. We generate captions for 9,646 videos: 7,985 from Charades \cite{charades} and 1,661 from AVCaps \cite{sudarsanam2025avcaps}. You can find the coverage of the dataset in the pie chart shown in the Figure \ref{fig:data_chart}.



\subsection{SFT with Phase-Switching}
\label{sec:sft}
We fine-tune SmolVLM2-500M via LoRA\cite{lora} in a single training run that internally switches between two phases, combining projector-only and full-adapter training without checkpoint restarts. Unless otherwise specified, we adopt the training configuration used in SmolVLM2\cite{smolvlm2}.

\paragraph{Phase 1: Visual Alignment (steps 0--$T_s$).}
During the first 25\% of training ($T_s = 0.25{\times}T_\text{total}$), only the SigLIP encoder and multimodal connector are trainable; LoRA adapters remain frozen. This aligns visual representations with the AD distribution before language model adaptation, preventing early overfitting to misaligned features.

\paragraph{Phase 2: Full Adaptation (steps $T_s$--$T_\text{total}$).}
A callback unfreezes all LoRA adapters and halves the vision encoder LR to $5{\times}10^{-6}$. We implement this via a custom \texttt{DifferentialLRTrainer} with three independent AdamW parameter groups.

\paragraph{Label Masking.}
Causal LM loss is computed over assistant response tokens only. Prompt tokens (system + user + image tokens) are masked to $-100$. Prompt length is determined by tokenising the prompt-only portion separately—avoiding error-prone token-ID-based detection.








\subsubsection{GRPO}
\label{sec:grpo}

Group Relative Policy Optimization\cite{grpo} is our primary RL method. For each prompt $x$, the policy samples $G{=}4$ completions $\{y_i\}$, scored by reward $R(x, y_i)$ normalised within the group:
\begin{equation}
    \hat{r}_i = \frac{R(x,y_i)-\mu_R}{\sigma_R};
\end{equation}
\begin{equation}
    \mu_R,\sigma_R = \operatorname{mean/std}_{j}\bigl(R(x,y_j)\bigr)
\end{equation}
The policy is updated to increase the probability of high-$\hat{r}$ completions.

\paragraph{Composite BLV Reward.}
$R(x,y)$ comprises six deterministic, rule-based components computable in $<$1\,ms per sample (Table\ref{tab:reward}).

\subsection{SFT-Patch Recovery}
\label{sec:sft_patch}

GRPO improves NAF but the RL optimization makes the policy focus on reward maximization, which degrades the Ambience sub-dimension of MCF (3.44$\to$2.94). To recover general descriptive quality, we apply a lightweight SFT-patch: 3 epochs on 50 curated Gemma captions, using the GRPO checkpoint as base with a fresh LoRA adapter \cite{hu2022lora} ($r{=}64$, $\alpha{=}128$, LR $5{\times}10^{-5}$, batch size 2, gradient clip 1.0). At inference, all three LoRA adapters (SFTv2, GRPO, SFT-patch) are merged sequentially: $\mathbf{W}' = \mathbf{W} + \mathbf{AB}$, yielding a single model with no additional inference overhead.

\begin{table}[h]
\centering
\small
\setlength{\tabcolsep}{3pt}
\begin{tabular}{@{}p{2.0cm}p{3.5cm}r@{}}
\toprule
\textbf{Component} & \textbf{Criterion} & \textbf{Score} \\
\midrule
\textsc{Caution} & Both ref.\ \& gen.\ contain it & $+0.40$ \\
 & Ref.\ has, gen.\ missing & $-0.30$ \\
 & Gen.\ has, ref.\ missing & $-0.10$ \\
\midrule
Directional & \{left, right, center, ahead, behind\} & $+0.08$/word \\
 & & (max $+0.24$) \\
\midrule
Distances & $\geq$1 metric distance match & $+0.20$ \\
\midrule
Structure & 2--4 sentences & $+0.16$ \\
 & $>$6 sentences & $-0.10$ \\
\midrule
Hallucination & Hedging phrases\textsuperscript{†} & $-0.20$ \\
\midrule
Env.\ ID & First sentence names scene\textsuperscript{‡} & $+0.10$ \\
\bottomrule
\end{tabular}
\caption{GRPO reward components. Range: $[-0.50,\,{+1.06}]$. \textsuperscript{†}\textit{``appears to be,''} \textit{``I cannot,''} etc. \textsuperscript{‡}14 scene-type terms.}
\label{tab:reward}
\end{table}

\subsection{Mobile Deployment}
\label{sec:deployment}

\paragraph{Model Preparation \& Conversion:}
All three LoRA adapters (SFT v2, GRPO, SFT-patch) are folded directly into the base model weights, producing a single unified HuggingFace checkpoint (969 MB, FP16). This merged model is then converted to the GGUF format via \texttt{llama.cpp}, a process that inherently separates the architecture into two distinct components for deployment: the language model backbone (783 MB, F16) and the vision projector (\texttt{mmproj.gguf}, 191 MB, F16).

\paragraph{Importance Matrix Calibration:}
To optimize the weights for our specific task, \texttt{llama-imatrix} is run on a calibration dataset of 500 BLV captions. This records which weight channels are most frequently activated, generating an importance matrix that guides non-uniform bit allocation during quantization.

\paragraph{Mixed-Precision Quantization Strategy:}
To compress the model while preserving its ability to generate accurate descriptions, we employed a mixed-precision quantization pipeline. The majority of the network was compressed using importance-weighted, non-linear 4-bit quantization (IQ4\_NL) \cite{malakhov2025deploying}. Guided by an importance matrix derived from our calibration dataset, this method maps weights down to an effective 4.5 bits by concentrating precision around the most frequently used values. However, applying uniform 4-bit compression degrades the model's positional reasoning. To mitigate this, we explicitly overrode the self-attention projections (q\_proj, k\_proj, v\_proj) to a higher 5-bit precision (Q5\_K). Protecting these specific attention mechanisms is critical for maintaining directional coherence when mapping visual tokens to text. This targeted override costs only a marginal increase in memory but substantially preserves output accuracy, yielding a highly optimized language backbone of 251 MB.
\paragraph{On-Device Inference:}
The two GGUF files (442\,MB total) are deployed on a Samsung Galaxy A55 (Exynos1480, 8\,GB RAM). \texttt{llama-mtmd-cli} is compiled natively in Termux. The vision projector encodes the input keyframe into token embeddings fed into the language model alongside the BLV navigation prompt.

\FloatBarrier
\section{Experiments}
\label{sec:results}
\paragraph{Experimental details.}
All GRPO experiments use the best SFT checkpoint as initialization and are trained with 4-bit NF4 quantization, bfloat16 compute, and a single 48\,GB RTX~A6000 GPU. We use group size $G{=}4$, a maximum generation length of 256 tokens, learning rate $5{\times}10^{-6}$, batch size 1 with gradient accumulation of 8, KL coefficient $\beta{=}0.1$, weight decay 0.01, and train for 2 epochs, corresponding to 9,076 optimization steps. All LLM-judge scoring (MCF and NAF) uses Qwen2.5-32B run locally. Prior to adoption, we manually reviewed a sample of judge outputs against human ratings to validate scoring reliability. In the experimental results, we use four training configurations:
Base (A), Base+SFT (B), Base+SFT+GRPO (C), and Base+SFT+GRPO+Patch (D).

\subsection{Main Results}
\paragraph{BLV improves OCR and TextVQA.}
Table~\ref{tab:ocr_vqa_comparison} shows that BLV-oriented fine-tuning improves not only accessibility-focused description, but also general text-centric visual reasoning. These results indicate that training on spatially grounded BLV descriptions strengthens the model's ability to parse visual text, spatial layout, and object-text relations.

\begin{table}
\centering
\small
\setlength{\tabcolsep}{4pt}
\resizebox{0.48\textwidth}{!}{ 
\begin{tabular}{@{}lcccc@{}}
\toprule
\textbf{Model} & \textbf{Params} & \textbf{OCR ANLS} & \textbf{VQA Accuracy} & \textbf{VQA ANLS} \\
\midrule
SmolVLM2-256M base & 256M & 16.49\% & 1.08\% & 3.09\% \\
SmolVLM2-500M base & 500M & 32.54\% & 37.75\% & 47.52\% \\
SmolVLM2-2.2B base & 2.2B & 25.44\% & 27.51\% & 29.64\% \\
LLaVA-1.5-7B & 7B & 28.78\% & 47.79\% & 62.96\% \\
PaliGemma-3B & 3B & 69.76\% & 76.09\% & 87.36\% \\
Qwen2-VL-2B & 2B & 86.10\% & 80.73\% & 89.24\% \\
Qwen2.5-VL-3B & 3B & 88.56\% & 81.12\% & 90.39\% \\
Qwen2-VL-7B & 7B & 87.23\% & 85.09\% & 92.89\% \\
\rowcolor{oursrow}
\textbf{Smol-VL-BLV (Ours)} & \textbf{500M} & \textbf{65.54\%} & \textbf{54.47\%} & \textbf{65.97\%} \\ \\
\bottomrule
\end{tabular}
}
\caption{Comparison of OCR and VQA performance across different vision-language models. Our 500M model substantially improves over the SmolVLM2-500M baseline and remains competitive with larger models.}
\label{tab:ocr_vqa_comparison}
\end{table}

\paragraph{Cross-Model Comparison on BLV Quality under Efficiency.} Table~\ref{tab:cross_model} compares Smol-VL-BLV with representative VLM baselines under BLV-oriented quality and computational cost. At the same 0.5B parameter scale and GFLOP budget as the SmolVLM2-500M baseline, our model improves Spatial score from 3.21 to 3.83 and Social score from 3.30 to 3.79. It also substantially improves general visual-text reasoning, increasing OCRBench ANLS from 32.5\% to 65.5\% and TextVQA accuracy from 37.8\% to 54.5\%. These results suggest that BLV-focused spatial post-training transfers beyond accessibility-specific captioning and strengthens OCR and VQA capabilities. Compared with larger VLMs, our model provides a favorable quality: compute trade-off. While Qwen2.5-VL-3B achieves higher OCR and VQA scores, it uses 6$\times$ more parameters. Smol-VL-BLV therefore offers strong BLV-oriented spatial and social understanding in a compact model, making it more suitable for resource-constrained BLV assistance.

\begin{table}
\centering
\small
\setlength{\tabcolsep}{3pt}
\resizebox{0.48\textwidth}{!}{ 
\begin{tabular}{@{}lrcccccc@{}}
\toprule
\textbf{Model} & \textbf{Params} & \textbf{Spatial} & \textbf{Social} & \textbf{OCR} & \textbf{VQA} & \textbf{GFLOPs} \\
 & & \multicolumn{2}{c}{\scriptsize(MCF sub-dims)} & & & \\
\midrule
PaliGemma2-3B & 3B & 1.15 & 1.02 & 7.8 & 14.3 & 10,543 \\
moondream2 & 1.8B & 1.00 & 1.00 & -- & -- & -- \\
SmolVLM2-2.2B & 2.2B & 3.49 & 3.45 & 25.4 & 27.5 & 125,406 \\
SmolVLM2-500M(base) & 0.5B & 3.21 & 3.30 & 32.5 & 37.8 & 34,855 \\
\rowcolor{oursrow}
\textbf{Smol-VL-BLV (Ours)} & \textbf{0.5B} & \textbf{3.83} & \textbf{3.79} & \textbf{65.5} & \textbf{54.5} & \textbf{34,855} \\
Qwen2.5-VL-3B & 3B & 5.43 & 3.51 & 88.6 & 81.1 & 14,280 \\
\bottomrule
\end{tabular}
}%
\caption{Overall comparative analysis ($n{=}47$, GPU server). Spatial and Social are MCF sub-dimension scores (1--10 scale). OCR = OCR-Bench ANLS (\%). VQA = TextVQA Accuracy (\%). ms = mean latency.}
\label{tab:cross_model}
\end{table}

\paragraph{Mobile Deployment Results.}
\label{sec:mobile_results}

Table~\ref{tab:deployment} reports on-device benchmarks across deployment iterations on the Samsung Galaxy A55 ($n{=}21$ keyframes for the final configuration).

\begin{table}
\centering
\small
\resizebox{0.48\textwidth}{!}{
\begin{tabular}{@{}lccccc@{}}
\toprule
\textbf{Metric} & \textbf{Paper Baseline} & \textbf{SFT} & \textbf{GRPO} & \textbf{GRPO + sft patch} \\
 \textbf{Quantization}& \textbf{Vivo Y27, INT8} & \textbf{A55, Q4\_K\_M } & \textbf{A55, Q4\_K\_M} & \textbf{A55, IQ4\_NL} \\
\midrule
Model (LM) & -- & 290 MB & 290 MB & 251 MB \\
Total on-device & -- & 481 MB & 481 MB & 442 MB \\
RAM load time & -- & 504 ms & 488 ms & 291 ms \\
TTFT (prompt eval) & -- & ${\sim}35$ s & 35.3 s avg & 25.1 s avg \\
Generation speed & 13.55 tok/s & 17.1 tok/s & 19.1 tok/s & 39.3 tok/s \\
Total per frame & 29.9 s & 44.2 s & 38.7 s & 27.1 s \\
Peak RAM & 761 MB & 619 MB & 985 MB & 780 MB \\
Model quality (MCF) & -- & 4.38 & 4.71  & same weights \\
\bottomrule
\end{tabular}
}
\caption{On-device performance comparison across model versions and hardware configurations}
\label{tab:deployment}
\end{table}

The IQ4\_NL quantization with importance-matrix calibration yields substantial gains over both the paper baseline and the initial Q4\_K\_M deployment. Generation speed more than doubles relative to Q4\_K\_M (+106\%), and at 39.3\,tok/s also surpasses the paper's reported 13.55\,tok/s by nearly 3x . TTFT and total per-frame latency fall by 29\% and 30\% respectively, while the model file shrinks 13\% and peak RAM drops 21\% — all relative to Q4\_K\_M. No crashes or out-of-memory errors were observed across all 21 evaluation frames.

For cross-device context, we evaluated the same 251 MB quantized model on a cloud T4 GPU, the A55 CPU, and an Apple Silicon Mac using Metal (Table~\ref{tab:crosshw}). Full generation completes in under 2.5 s on both the T4 and Mac, compared with 27.1 s on the A55. The A55 uses CPU-only inference due to unavailable hardware acceleration support (Appendix~\ref{app:npu}). Thus, its 27.1 s latency reflects the deployment limitation rather than the model's potential with GPU/NPU acceleration.

\begin{table}[t]
\centering
\small
\setlength{\tabcolsep}{4pt}
\resizebox{0.48\textwidth}{!}{
\begin{tabular}{lccc}
\hline
Metric & Cloud (T4) & A55 (CPU) & Mac (Metal) \\
\hline
TTFT / Prefill & 1.39 s & 25.1 s & 1.95 s \\
Generation Time & 0.31 s & 1.4 s & 0.46 s \\
Total Latency & 1.70 s & 27.1 s & 2.41 s \\
New Tokens & 58 & 60 & 61 \\
Gen. Speed (tok/s) & 187.42 & 42.9 & 132.6 \\
Latency/Token (ms) & 5.34 & 23.3 & 7.54 \\
\hline
\end{tabular}
}
\caption{Cross-hardware latency for the same quantized model. The A55's disproportionate latency reflects CPU-only fallback from missing NPU/GPU driver support (Appendix~\ref{app:npu}), not a property of the model itself.}
\label{tab:crosshw}
\end{table}
\paragraph{Human Evaluation}
Twenty sighted crowd work raters evaluated five real time demo videos across different environments (obstacles, floor objects, doors). Each video was rated from 0–5 on five criteria: Navigation Relevance, CAUTION/Alert Accuracy, Spatial and Directional Precision, Response Conciseness, and Overall Trustworthiness.  Navigation Relevance measures whether the description contains directly actionable spatial content: obstacles, directions, and distances. CAUTION/Alert Accuracy evaluates whether hazards are correctly identified. Spatial and Directional Precision captures whether grounded spatial cues, such as left/right orientation and estimated distances, are present. Response Concise reflects whether the description is delivered within two sentences, satisfying the low-latency constraint of mobile deployment.

Mean scores across evaluators were: Navigation Relevance (4.0), CAUTION/Alert Accuracy (3.5), Spatial and Directional Precision (4.5), Response Conciseness (4.5), and Overall Trustworthiness (4.0), yielding an overall average of 4.1 out of 5. The high scores on spatial precision and conciseness confirm the model's suitability as a navigation aid, while the lower hazard accuracy score reflects the inherent difficulty of conveying hazard severity through language alone.





\begin{table}
\centering
\small
\setlength{\tabcolsep}{4pt}
\resizebox{0.48\textwidth}{!}{%
\begin{tabular}{@{}lccc>{\columncolor{oursrow}}c>{\columncolor{oursrow}}c@{}}
\toprule
\textbf{Metric} & \textbf{A}(Base) & \textbf{B}(SFT) & \textbf{C}(+GRPO) & \textbf{D}(+Patch) & \textbf{$\Delta$} \\
\midrule
BLEU-1  & 19.02  & 18.08  & 22.55  & \textbf{45.57}  & +26.55 \\
BLEU-4  & 0.74   & 1.66   & 1.97   & \textbf{13.53}  & +12.79 \\
ROUGE-L & 10.74  & 17.66  & 17.93  & \textbf{30.49}  & +19.75 \\
METEOR  & 13.76  & 15.65  & 17.68  & \textbf{35.99}  & +22.23 \\
CIDEr   & 0.0002 & 0.0021 & 0.0018 & \textbf{0.0112} & +0.011 \\
\bottomrule
\end{tabular}}
\caption{NLP metrics on the 458-sample final evaluation set. Condition~D achieves large gains across all metrics.}
\label{tab:main_nlp}
\end{table}


\begin{table}
\centering
\small
\setlength{\tabcolsep}{4pt}
\resizebox{0.48\textwidth}{!}{%
\begin{tabular}{@{}lcccc@{}}
\toprule
\textbf{Metric} & \textbf{A} (Base) & \textbf{B} (SFT) & \textbf{C} (+GRPO) & \textbf{D} (+Patch) \\
\midrule
\multicolumn{5}{l}{\textit{BLV Coverage (\%)}} \\
Spatial   & 79 & 81 & 97 & \cellcolor{oursrow}\textbf{100} \\
Social    & 94 & 94 & 98 & \cellcolor{oursrow}\textbf{100} \\
Action    & 68 & 79 & \cellcolor{oursrow}\textbf{88} & 62 \\
Ambience  & 76 & 91 & 91 & \cellcolor{oursrow}\textbf{98} \\
BLV Mean  & 79 & 86 & \cellcolor{oursrow}\textbf{93} & 90 \\
\midrule
\multicolumn{5}{l}{\textit{Navigational Coverage (\%)}} \\
Obstacles       & 77 & 85 & 92 & \cellcolor{oursrow}\textbf{96} \\
Step changes    & 4  & 7  & 7  & \cellcolor{oursrow}\textbf{5} \\
Directions      & 29 & 51 & 88 & \cellcolor{oursrow}\textbf{100} \\
Moving hazards  & 16 & 21 & 17 & \cellcolor{oursrow}\textbf{66} \\
Distances       & 45 & 20 & 22 & \cellcolor{oursrow}\textbf{100} \\
Nav.\ Mean      & 34 & 37 & 45 & \cellcolor{oursrow}\textbf{73} \\
\bottomrule
\end{tabular}
}
\caption{BLV keyword coverage (458 samples). Condition~D achieves 100\% coverage on spatial orientation, directions, and distances.}
\label{tab:main_blv}
\end{table}

\begin{table}[t!]
\centering
\small
\setlength{\tabcolsep}{3pt}
\begin{tabular}{@{}lcccc@{}}
\toprule
\textbf{Dimension} & \textbf{A} (Base) & \textbf{B} (SFT) & \textbf{C} (+GRPO)& \textbf{D} (+Patch)\\
\midrule
\multicolumn{5}{l}{\textit{LLM Judge (1--10 scale)}} \\
MCF Score & 1.88 & 2.26 & 2.39 & \cellcolor{oursrow}\textbf{2.94} \\
NAF Score & 1.84 & 2.44 & 2.35 & \cellcolor{oursrow}\textbf{3.57} \\
\midrule
\multicolumn{5}{l}{\textit{MCF Sub-Dimensions}} \\
Spatial orient.   & 1.21 & 1.71 & 1.85 & \cellcolor{oursrow}\textbf{3.23} \\
Social interact.  & 1.47 & 1.62 & 1.82 & \cellcolor{oursrow}\textbf{2.60} \\
Action events     & 2.16 & 2.36 & 2.47 & \cellcolor{oursrow}\textbf{3.00} \\
Ambience          & 2.70 & 3.37 & \cellcolor{oursrow}\textbf{3.44} & 2.94 \\
\midrule
\multicolumn{5}{l}{\textit{NAF Sub-Dimensions}} \\
Descriptiveness & 1.46 & 1.85 & 1.92 & \cellcolor{oursrow}\textbf{2.87} \\
Objectivity     & 2.71 & 3.77 & 3.38 & \cellcolor{oursrow}\textbf{5.38} \\
Accuracy        & 1.09 & 1.59 & 1.55 & \cellcolor{oursrow}\textbf{2.42} \\
Clarity         & 2.09 & 2.55 & 2.54 & \cellcolor{oursrow}\textbf{3.63} \\
\bottomrule
\end{tabular}
\caption{LLM judge scores and sub-dimension breakdown (1--10, 458 samples). Condition~D achieves the highest scores on 7 of 8 dimensions, with GRPO (C) leading only on ambience.}
\label{tab:judge_subdim}
\end{table}

\vspace{3mm}
\begin{table}[t!]
\centering
\small
\begin{tabular}{@{}lccc@{}}
\toprule
\textbf{Method} & \textbf{MCF} & \textbf{NAF} & \textbf{Overall} \\
\midrule
Base (zero-shot) & 3.90 & 3.98 & 3.92 \\
SFT v2 & \textbf{4.38} & 3.98 & 3.97 \\
\midrule
DPO & 3.89 & 3.98 & 3.92 \\
RLAIF-V DPO & 3.88 & 3.96 & 3.91 \\
\rowcolor{oursrow}
GRPO & 4.18 & \textbf{4.20} & \textbf{4.19} \\
\bottomrule
\end{tabular}
\caption{Preference alignment comparison on the balanced evaluation set. Base, SFT v2, DPO, and RLAIF-V DPO are evaluated on 469 samples using a 1--5 judge scale. The DPO and RLAIF-V DPO fall below SFT v2 on MCF.}
\label{tab:pref_compare}
\end{table}






\paragraph{Qualitative Results}
\label{sec:qualitative}

\begin{figure*}[ht]
    \centering
    \includegraphics[width=\linewidth]{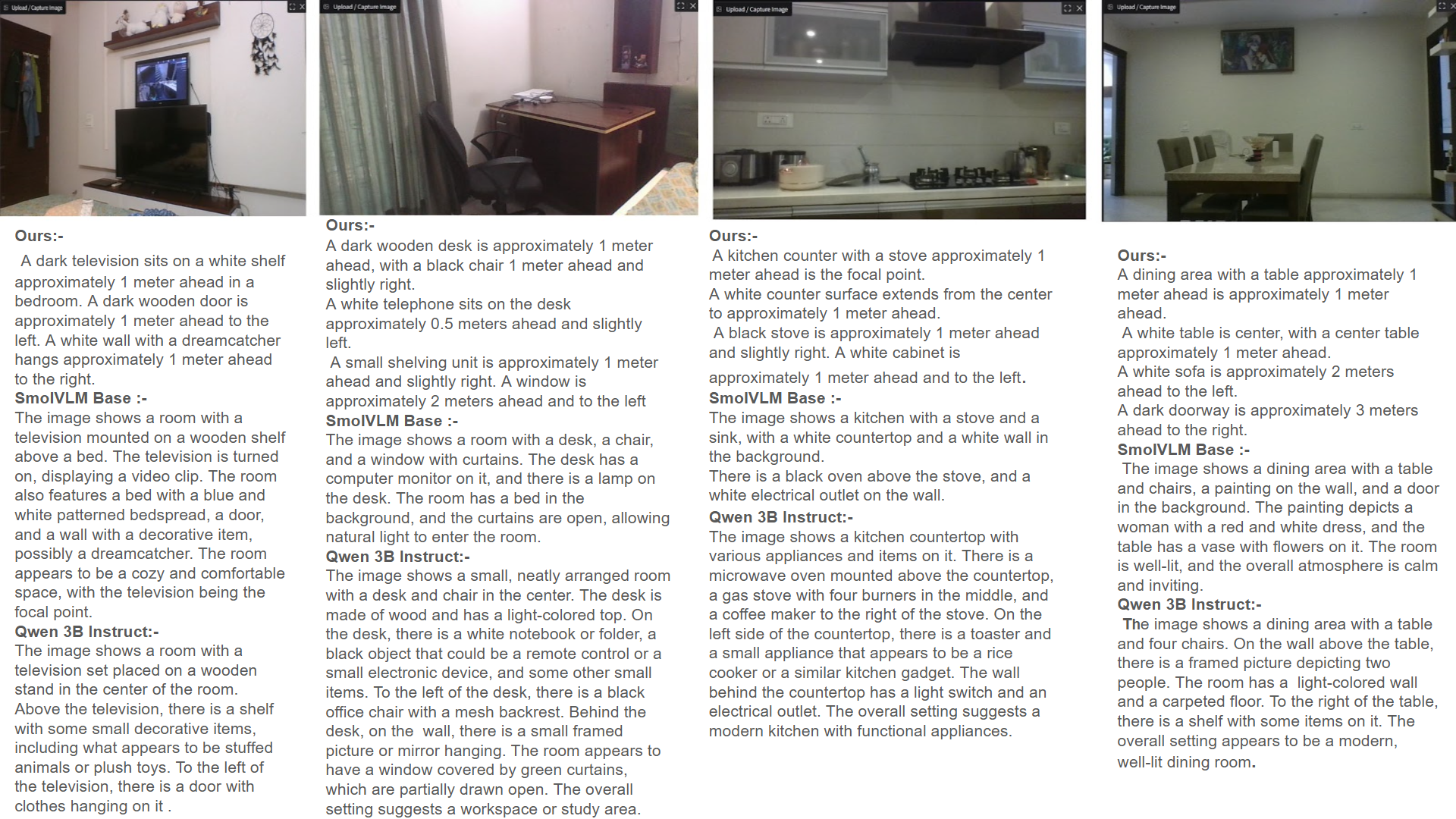}
    \caption{Qualitative comparison of scene descriptions generated by our model, SmolVLM-500M (Base), and Qwen-3B across four indoor scenes. Our model produces concise, spatially grounded, navigation-focused descriptions, while base models generate generic, observation-style captions without actionable spatial information.}
    \label{fig:quality_comp}
\end{figure*}

In the Figure~\ref{fig:quality_comp} illustrates the progressive improvement from base to fine-tuned outputs on a representative indoor scene. Our model's results show that compact and BLV are relevant in complex scenes.

\subsection{Discussion}
Some discussions on emergent capabilities after spatial grounding SFT and GRPO finetuning of SmolVLM-500M : \textbf{(1)} The spatially grounded question-answer pairs generated by Gemma 4, included the state of the environment, nearby hazards, directional cues, person's position, color of cloths, and their movements. Spatial awareness needs the model to specialise in coordinate mapping, relative positioning and bounding logic. \textbf{(2)} Using a very good teacher (Gemma 4) to generate the BLV focused description having spatial awareness, forces the SmolVLM to build spatial maps for BLV navigation. \textbf{(3)} This structural alignment makes it much sharper at parsing the geometric boundaries of characters and text strings, so there is a significant improvement in the OCR performance of the finetuned model over the baseline SmolVLM. \textbf{(4)} Small models have a tendency to copy the style and rigor of their training data. The injection of high-quality synthetic data from Gemma 4 makes the SmolVLM adopt a better reasoning strategy, which can be clearly inferred from Table 8, resulting in 44 \% improvement from baseline. \textbf{(5)} Early GRPO experiments used a few-shot BLV examples in the prompt, but this led the model to imitate their structure instead of grounding outputs in the scene. We therefore switched to zero-shot prompting during GRPO and used the later SFT-patch stage to improve descriptive quality. We also note that CIDEr scores remain low in absolute terms across all conditions (Table~\ref{tab:main_nlp}) this is expected under single reference evaluation, since CIDEr's n-gram/TF-IDF overlap is typically computed against multiple human references, making exact overlap structurally rare here with only one teacher generated reference per video. The relative improvement across conditions remains the meaningful signal.


\subsection{Ablation Studies}
\label{sec:ablations}

This design allows us to isolate the contribution of each stage: supervised distillation for output structure, GRPO for BLV-specific preference alignment, and the patch-based SFT stage for recovering descriptive quality and improving spatial grounding.

\paragraph{Stages of post-training.}
The cumulative ablation shows that each post-training stage contributes a distinct improvement. SFT first establishes the basic audio-description structure, improving general description quality and BLV coverage, with ROUGE-L increasing from 10.74 to 17.66, METEOR from 13.76 to 15.65, and BLV Mean from 79\% to 86\% (Tables~\ref{tab:main_nlp} and~\ref{tab:main_blv}). GRPO then strengthens BLV-specific alignment by increasing BLV Mean to 93\%, improving directional coverage from 51\% to 88\%, and raising obstacle coverage from 85\% to 92\%, showing that the online reward encourages navigationally useful descriptions. Finally, the SFT-patch stage produces the strongest overall gains: condition D achieves the best NLP scores and improves spatial coverage to 100\%, direction coverage to 100\%, distance coverage to 100\%, and moving-hazard coverage to 66\%. These results indicate that SFT transfers the teacher's descriptive structure, GRPO reinforces BLV-oriented spatial and hazard cues, and the patch stage provides localized visual grounding for precise metric and directional descriptions.
\paragraph{LLM Judge: Sub-Dimension Analysis}
\label{sec:judge_detail}

Table~\ref{tab:judge_subdim} breaks down the LLM judge scores across all eight dimensions for the four primary conditions.
The most striking improvement is in \textbf{objectivity}: from 2.71 (Base) to 5.38 (D), a +2.67 gain. This indicates that the pipeline successfully eliminates subjective hedging (e.g., \textit{``appears to be,'' ``seems like''}) and produces factual, AD-compliant descriptions. \textbf{Spatial orientation} shows the second-largest gain (+2.02), directly reflecting the GRPO reward function's emphasis on directional language and metric distances.


\paragraph{Comparison with other preference-alignment methods.}
Table~\ref{tab:pref_compare} compares GRPO with offline preference-alignment methods. 
DPO and RLAIF-V do not improve over SFT, with MCF scores of 3.89 and 3.88 compared with 4.38 for SFT v2. 
In contrast, GRPO achieves stronger NAF performance, reaching 4.20 after rescaling. 
This suggests that offline preference pairs may capture general caption quality but are less effective at enforcing BLV-specific navigational behavior. 
The online GRPO reward is more suitable because it directly optimizes directional language, distance references, and hazard awareness.

\section{Conclusion}

This work demonstrates that a $<$1B VLM which is targeted to run on-device with resource constraints, can be effectively specialized for low vision-compliant scene narration for navigational assistance. Starting from baseline, our distillation pipeline transfers spatial awareness and safety from a larger teacher, resulting in a model that outperforms even larger models on custom and benchmark evaluation criteria.
We observed SFT alone establishes a strong general baseline, but it is GRPO that drives the gains in navigational content that define BLV utility. GRPO is not just a contributor to the RL stage, it is the underlying driver of the best overall model's superiority. The on-device benchmarking results on a mid-range Android handset makes it practically deployable and meaningfully superior to larger baselines.

\section{Future Work}
\label{sec:future-work}

There are several directions that we would like to explore in future work. We plan to collect more training data covering stairs, curbs, elevation changes, and moving hazards to address the gaps observed in Table~\ref{tab:main_nlp}. We also plan to introduce a dedicated step-change reward during the GRPO stage to encourage the model to learn this behavior directly. Further, we will evaluate the system in challenging environments such as crowded crosswalks and busy intersections to test its robustness beyond the current dataset. We also plan to improve our evaluation by developing a geometry based spatial correctness metric, reducing our reliance on keyword matching and LLM judge scores. Finally, we will add a separate factual accuracy check for incorrect directions, false hazard claims, and fabricated distances to better assess the system's reliability and safety.

\section*{Limitations}
This work primarily targets low-vision users; while we attempted to extend its utility to fully blind users, the system does not yet meet the reliability required for safe independent blind navigation. The dataset used is limited to some indoor and outdoor scenes, which can be improved to more difficult navigation related domains such as road and crowded scenes. The evaluation can be made by BLV users.
Coverage gaps remain for step-change (5\%) and moving-hazard (66\%) detection Table ~\ref{tab:main_blv}. We attribute this primarily to under representation of these scenarios in our training data Figure~\ref{fig:data_chart} rather than a limitation of the GRPO reward function itself. Stairs, curbs, elevation changes, and moving hazards are comparatively rare events in these source datasets, giving the model few positive examples to learn from.
Distances generated in the teacher descriptions are approximate visual judgments derived from the teacher model rather than sensor measured values.

\section*{Ethics Statement}
Assistive AI systems for low-vision people raise issues of privacy, consent for bystanders, overreliance, safety, accessibility equity, and representation of disability communities. User data should be collected with informed consent and protected carefully.

\bibliography{main}

\appendix

\section{Deployment Architecture \& Optimization}
\label{app:deployment}
 
\subsection{Model Architecture}
\label{app:model-arch}
 
Our deployed model is SmolVLM2-500M-Video-Instruct with a three-stage LoRA training
lineage. Each stage added a LoRA adapter on top of the previous, and all three adapters
were merged into the base model weights before export, producing a single
HuggingFace model folder with no separate adapter files.
 
\subsubsection{Neural Network Breakdown}
 
\begin{table}[H]
\centering
\small
\setlength{\tabcolsep}{2pt}
\begin{tabularx}{\columnwidth}{>{\raggedright\arraybackslash}p{2.0cm}
                                >{\raggedright\arraybackslash}X
                                >{\centering\arraybackslash}p{1.4cm}
                                >{\raggedright\arraybackslash}p{2.4cm}}
\toprule
\textbf{Component} & \textbf{Architecture Detail} & \textbf{Params} & \textbf{Role} \\
\midrule
Vision Encoder
  & SigLIP, ViT-style, Patch/14, 384$\times$384 input
  & $\sim$86.4M
  & Pixels $\to$ patch embeddings \\[2pt]
Vision Projector
  & Linear projection + LayerNorm
  & $\sim$11.8M
  & Vision space $\to$ LM tokens \\[2pt]
Language Model
  & LLaMA arch, 32 blocks, ctx\,8192, vocab\,49280
  & $\sim$409.3M
  & Text generation \\[2pt]

Full model
  & \texttt{blv\_final/} on server
  & $\sim$507.5M
  & Single artifact \\
\bottomrule
\end{tabularx}
\caption{Neural network component breakdown of the deployed model.}
\label{tab:nn-breakdown}
\end{table}
 
\subsubsection{Why the Model Splits into Two GGUF Files}
 The model produces two GGUF files totaling 442\,MB, rather than the single 103\,MB
file used by the unmodified paper baseline. When converting any VLM via
\texttt{convert\_hf\_to\_gguf.py}, the \texttt{llama.cpp} script always separates
the vision projector (\texttt{mmproj}) from the language model backbone by design,
since the two components run sequentially and can be quantized independently. The
paper baseline used the original unmodified SmolVLM2-500M (no LoRA, no fine-tuning),
whose architecture the conversion script packed into one file. Once LoRA adapters are
merged and the modified weights are exported, the script treats them differently. At
runtime, \texttt{llama-mtmd-cli} loads both files into RAM, runs \texttt{mmproj} to
encode the image into vision token embeddings, feeds those into the LM's token stream,
and the LM generates text. From the user's perspective, the system behaves as a single
model---two files is purely a storage artifact.

\subsubsection{Full Deployment Pipeline}
 
\begin{figure}[H]
\centering
\begin{adjustbox}{max width=\columnwidth}
\begin{minipage}{0.98\columnwidth}
\ttfamily\small\linespread{1.1}\selectfont
\begin{tabbing}
\textbf{blv\_final/} \hspace{2pt} (merged HF model, float16, \textasciitilde969\,MB)\\
\hspace{4pt} SFT v2 + GRPO + SFT Patch v2 LoRAs -- all folded in\\[2pt]
\hspace{40pt} $\downarrow$ \hspace{4pt} \textrm{Step 1:} \texttt{convert\_hf\_to\_gguf.py}\\[2pt]
\textbf{blv\_final\_f16.gguf} \hspace{4pt} (GGUF, F16, 783\,MB)\\
\textbf{sft\_patch\_mmproj.gguf} \hspace{4pt} (vision projector, extracted separately)\\[2pt]
\hspace{40pt} $\downarrow$ \hspace{4pt} \textrm{Step 2:} \texttt{llama-imatrix} \textrm{ (BLV calibration, \textasciitilde30\,min)}\\
\hspace{40pt} $\downarrow$ \hspace{4pt} \textrm{Step 3:} \texttt{llama-quantize} \textrm{ IQ4\_NL + imatrix + Q5\_K attn}\\[2pt]
\textbf{blv\_final\_iq4nl\_best.gguf} \hspace{4pt} 251\,MB \hspace{4pt} (LM backbone, IQ4\_NL)\\
\textbf{sft\_patch\_mmproj.gguf} \hspace{4pt} 191\,MB \hspace{4pt} (vision projector, F16)\\
\hspace{4pt} Total on-device: \textbf{442\,MB}\\[2pt]
\hspace{40pt} $\downarrow$ \hspace{4pt} \textrm{ADB push / USB transfer}\\[2pt]
\textbf{Samsung Galaxy A55} \hspace{4pt} (Exynos\,1480, 8\,GB RAM, Android\,16)\\
\hspace{4pt} Termux + \texttt{llama-mtmd-cli} \hspace{4pt} (CPU-only, 4 threads)\\
\hspace{8pt} \textbullet{} mmproj encodes frame.jpg $\to$ vision token embeddings\\
\hspace{8pt} \textbullet{} LM receives [vision tokens] + [BLV navigation prompt]\\
\hspace{8pt} \textbullet{} LM generates BLV description ($\leq$60 tokens, greedy)\\
\hspace{8pt} \textbullet{} \texttt{termux-tts-speak} reads output aloud
\end{tabbing}
\end{minipage}
\end{adjustbox}
\caption{End-to-end deployment pipeline from merged HuggingFace model to
on-device inference on Samsung Galaxy A55.}
\label{fig:deploy-pipeline}
\end{figure}
 
\subsection{Quantization}
\label{app:quantization}
 
The goal of quantization is to shrink the model from its training-time float16
representation (783\,MB) into something a phone CPU can load and run at
acceptable speed. We explored several approaches before settling on IQ4\_NL.
 
\subsubsection{Step-by-Step Procedure}
 
\paragraph{Step 1 --- Merge LoRA into base weights.}
After training, three LoRA adapters are stored as small delta matrices (each
32--77\,MB). These are added into the base model weights via
$W_{\text{merged}} = W_{\text{base}} + (A \times B)$, where $A$ and $B$ are the
low-rank matrices, producing \texttt{blv\_final/}. The merge takes $\sim$3\,minutes
on CPU with no GPU required. \texttt{llama.cpp}'s converter requires complete weight
matrices per layer and does not understand HuggingFace adapter files.
 
\paragraph{Step 2 --- Convert to GGUF (F16).}
\texttt{convert\_hf\_to\_gguf.py} reads the merged HuggingFace model and writes it
to GGUF format---a single binary packaging all weights plus model metadata
(architecture, vocabulary, context length). Weights remain float16 at this stage.
Output: \texttt{blv\_final\_f16.gguf} (783\,MB). The mmproj is extracted as a
separate file automatically. A known issue: the script triggers a
\texttt{torchvision} circular import through \texttt{transformers\,5.5.1}; fixed
with a \texttt{ModuleSpec} mock that registers a dummy \texttt{torchvision} module
before conversion runs.
 
\paragraph{Step 3 --- Generate importance matrix (imatrix).}
\texttt{llama-imatrix} is run against the F16 GGUF using a calibration dataset of
$\sim$500 BLV captions from training data, recording which weight channels were most
frequently activated. Standard quantization compresses all weights equally; the
imatrix directs the quantizer to allocate more bits to high-salience channels.
Runtime: $\sim$30\,minutes (one-time cost). Output: \texttt{blv\_imatrix.dat}.
 
\paragraph{Step 4 --- Quantize to IQ4\_NL with mixed precision.}
\texttt{llama-quantize} takes the F16 GGUF and imatrix file and produces the final
model using IQ4\_NL (importance-weighted, non-linear 4-bit), with a precision
override: attention layers (\texttt{q\_proj}, \texttt{k\_proj}, \texttt{v\_proj})
are quantized at Q5\_K because they directly control spatial and directional focus.
IQ4\_NL uses a non-linear mapping (float16 $\to$ 4.5\,bits effective), concentrating
precision around frequently-used values. The Q5\_K override costs $\sim$15\,MB but
improves BLV description coherence. Output: \texttt{blv\_final\_iq4nl\_best.gguf}
(251\,MB), 13\% smaller than the Q4\_K\_M baseline (290\,MB).
 
\paragraph{Step 5 --- Transfer and run on Android.}
The two GGUF files were transferred via USB using ADB. \texttt{llama-mtmd-cli} was
compiled directly on-device ($\sim$30--40\,min, one-time). Inference command:
{\small\texttt{llama-mtmd-cli -m blv\_final\_iq4nl\_best.gguf -{}-mmproj
sft\_patch\_mmproj.gguf -{}-image frame.jpg -p [BLV\_PROMPT] -n 60 -{}-temp 0.0 -t 4}}.
Output is read aloud via \texttt{termux-tts-speak}.
 
\subsubsection{Quantization Methods Explored}
 
\begin{table}[H]
\centering
\small
\setlength{\tabcolsep}{4pt}
\begin{tabularx}{\columnwidth}{>{\raggedright\arraybackslash}p{1.5cm}
                                >{\raggedright\arraybackslash}X
                                >{\raggedright\arraybackslash}X}
\toprule
\textbf{Method} & \textbf{What We Did} & \textbf{Outcome} \\
\midrule
Q4\_K\_M
  & Standard 4-bit K-quant; Apr\,25 \& May\,9 benchmarks.
  & 290\,MB; TTFT $\sim$35\,s; 17--19\,tok/s. Baseline for comparisons. \\[3pt]
Q5\_K\_M
  & 5-bit K-quant; tested spatial-language coherence gain.
  & $\sim$340\,MB. RSS $>$1050\,MB with mmproj; quality gain marginal. \\[3pt]
Q8\_0
  & Closest-to-F16 quality; evaluated theoretically.
  & 783\,MB; rejected---total RAM exceeds 1\,GB plus mmproj plus OS. \\[3pt]
IQ4\_XS
  & Importance-weighted 4-bit extra-small.
  & $\sim$230\,MB; noticeably worse coherence on BLV spatial descriptions. \\[3pt]
\textbf{IQ4\_NL} \newline \textbf{(selected)}
  & imatrix from BLV captions; Q5\_K on attn layers.
  & \textbf{251\,MB; TTFT 25.1\,s; 39.3\,tok/s.} Final deployed model. \\[3pt]
Q3 \& below
  & IQ3, Q3, Q2 variants; reviewed llama.cpp docs.
  & Rejected. Sub-4-bit renders small VLMs ($<$1B) incoherent (SPEED-Q). \\[3pt]
AWQ
  & Activation-aware quant via \texttt{autoawq}.
  & No support for SmolVLM2's SigLIP+LLaMA hybrid; \texttt{autoawq} targets LLaVA/Qwen-VL only. \\[3pt]
GPTQ
  & Hessian-based PTQ; calibration on BLV data.
  & Same blocker as AWQ. imatrix in IQ4\_NL is conceptually equivalent and works natively in \texttt{llama.cpp}. \\[3pt]
QAT
  & Re-train with fake quant ops in forward pass.
  & Requires full SFT+GRPO+SFT\_Patch rerun; \texttt{bitsandbytes} QAT experimental for VLMs. Future work. \\[3pt]
BitNet 1.58b
  & Ternary weights; integer-only arithmetic.
  & Inapplicable post-hoc; requires training from scratch with ternary constraints. \\
\bottomrule
\end{tabularx}
\caption{Quantization methods explored and outcomes.}
\label{tab:quant-methods}
\end{table}
 
\subsection{NPU Acceleration: Approaches Explored}
\label{app:npu}
 
TTFT is CPU-bound at $\sim$25\,s. The Exynos\,1480 in the A55 includes a dedicated
NPU. We investigated several acceleration routes.
 
\paragraph{Android NNAPI.}
NNAPI routes neural network computations to device accelerators (NPU/DSP/GPU).
\texttt{llama.cpp}'s Android build does not expose an NNAPI execution provider---its
GPU path targets OpenCL and Vulkan. Samsung's Exynos NNAPI driver also frequently
falls back to GPU for unsupported ops. A proper NNAPI pipeline would further require
a full Android app (Java/Kotlin + NDK), incompatible with our Termux CLI workflow.
 
\paragraph{ONNX Runtime Mobile + NNAPI.}
HuggingFace \texttt{optimum-cli export onnx} does not support SmolVLM2: its
SigLIP encoder combined with the pixel\_values mid-sequence injection has no
registered export class in \texttt{optimum}'s \texttt{TasksManager}. The exporter
raised \texttt{KeyError: 'smolvlm2'} immediately. A custom ONNX exporter would need
to implement the full SigLIP + token-injection logic as ONNX-compatible ops---not
feasible within the timeline.
 
\paragraph{MLC-LLM.}
MLC-LLM supports LLaMA, Mistral, and Qwen; SmolVLM2 is not included. Its Android
hardware acceleration path targets Snapdragon via Qualcomm ADRENO/Hexagon. On
Exynos\,1480, the app falls back entirely to CPU---NPU acceleration is unavailable
for non-Snapdragon devices.
 
\paragraph{Vulkan / OpenCL (\texttt{llama.cpp}).}
Building \texttt{llama.cpp} with \texttt{-DGGML\_VULKAN=ON} in Termux failed: the
Vulkan headers were found but the vendor-specific Mali ICD loader (needed to route
calls to hardware) was not. The CLBLAST/OpenCL path had the same issue---the Exynos
Mali OpenCL 2.0 driver resides in Samsung's proprietary partition, inaccessible from
Termux.
 
\paragraph{Pocketpal AI.}
This consumer GGUF inference app did not support vision (\texttt{mmproj}) at time of
testing. A later version requires LLaVA-style mmproj naming conventions, which our
export does not match. The GPU toggle reports ``No GPU backend available'' on A55
(same Vulkan blocker as above).
 
\paragraph{Samsung ONE SDK.}
Direct Exynos NPU programming via Samsung's proprietary ON-device Neural Engine SDK
requires NDA registration; access was not granted in time. The SDK also requires
models in \texttt{.nnc} format, produced from TFLite or ONNX---both of which were
blocked for SmolVLM2 (see above).
 
\subsection{Benchmark Results}
\label{app:benchmarks}
 
\subsubsection{IQ4\_NL Benchmark --- Samsung A55 ($n{=}21$ frames)}
 
\noindent Model: \texttt{blv\_final\_iq4nl\_best.gguf} (251\,MB) +
\texttt{sft\_patch\_mmproj.gguf} (191\,MB) = 442\,MB. Termux, 4 threads,
CPU-only. Date: May 24, 2026.
 
\begin{table}[H]
\centering
\small
\begin{tabular}{lccc}
\toprule
\textbf{Metric} & \textbf{Min} & \textbf{Avg} & \textbf{Max} \\
\midrule
Model load into RAM      & 272\,ms & 291\,ms & 321\,ms \\
Prompt eval (TTFT)       & 24.1\,s & 25.1\,s & 28.2\,s \\
Generation               & 1.0\,s  & 1.4\,s  & 1.7\,s  \\
Total per frame          & 26.0\,s & 27.1\,s & 30.5\,s \\
Prompt speed (tok/s)     & 32.7    & 37.0    & 38.2    \\
Generation speed (tok/s) & 34.2    & 39.3    & 43.2    \\
Peak RSS                 & ---     & $\sim$780\,MB & 1004\,MB \\
\bottomrule
\end{tabular}
\caption{Per-frame benchmarks for the final IQ4\_NL deployment.
2 of 21 frames showed thermal throttling; 19 frames sustained normal speed.
No crashes or OOM events.}
\label{tab:bench-iq4nl}
\end{table}
 
\subsubsection{Deployment History --- All Stages}
 
\noindent\textit{Note on TTFT.} Early benchmarks reported 504\,ms as TTFT; that
was the RAM load time. True TTFT---processing image and prompt tokens---is
$\sim$35\,s for Q4\_K\_M and $\sim$25\,s for IQ4\_NL.
 
\begin{table}[H]
\centering
\small
\begin{adjustbox}{max width=\columnwidth}
\setlength{\tabcolsep}{5pt}
\begin{tabular}{lcccc}
\toprule
\textbf{Metric} & \textbf{Paper Baseline} & \textbf{SFT} & \textbf{GRPO} & \textbf{GRPO + sft patch} \\
 \textbf{Quantization}& \textbf{Vivo Y27, INT8} & \textbf{A55, Q4\_K\_M } & \textbf{A55, Q4\_K\_M} & \textbf{A55, IQ4\_NL} \\
\midrule
Model (LM) & -- & 290 MB & 290 MB & 251 MB \\
Total on-device & -- & 481 MB & 481 MB & 442 MB \\
RAM load time & -- & 504 ms & 488 ms & 291 ms \\
TTFT (prompt eval) & -- & ${\sim}35$ s & 35.3 s avg & 25.1 s avg \\
Generation speed & 13.55 tok/s & 17.1 tok/s & 19.1 tok/s & 39.3 tok/s \\
Total per frame & 29.9 s & 44.2 s & 38.7 s & 27.1 s \\
Peak RAM & 761 MB & 619 MB & 985 MB & 780 MB \\
Model quality (MCF) & -- & 4.38 & 4.71  & same weights \\
\bottomrule
\end{tabular}
\end{adjustbox}
\caption{Deployment benchmark history across all stages.
$^*$Paper's 29.9\,s is for text-only inference (no vision tokens); the same paper
reports 60--83\,s for full multimodal inference with images.
$^\dagger$Paper baseline TTFT likely includes image encoding overhead in their
unified model and is not directly comparable to our split pipeline.}
\label{tab:bench-history}
\end{table}
 
\subsubsection{IQ4\_NL vs.\ Q4\_K\_M --- Summary}
 
\begin{table}[H]
\centering
\small
\begin{tabular}{lccc}
\toprule
\textbf{Metric} & \textbf{Q4\_K\_M (May\,9)} & \textbf{IQ4\_NL (May\,24)} & \textbf{$\Delta$} \\
\midrule
Total per frame          & 38.7\,s       & 27.1\,s        & $-$30\% \\
TTFT                     & 35.3\,s       & 25.1\,s        & $-$29\% \\
Generation speed         & 19.1\,tok/s   & 39.3\,tok/s    & $+$106\% \\
LM file size             & 290\,MB       & 251\,MB        & $-$13\% \\
Total on-device          & 481\,MB       & 442\,MB        & $-$8\%  \\
Peak RAM                 & 985\,MB       & $\sim$780\,MB  & $-$21\% \\
RAM load time            & 488\,ms       & 291\,ms        & $-$40\% \\
\bottomrule
\end{tabular}
\caption{IQ4\_NL vs.\ Q4\_K\_M improvement summary.}
\label{tab:bench-comparison}
\end{table}

\end{document}